\documentclass[letterpaper]{article} 
\usepackage{aaai2026}  
\usepackage{times}  
\usepackage{helvet}  
\usepackage{courier}  
\usepackage[hyphens]{url}  
\usepackage{graphicx} 
\usepackage{natbib}  
\usepackage{caption} 
\usepackage{algorithm}
\usepackage{algorithmic}

\usepackage{newfloat}
\usepackage{listings}
\DeclareCaptionStyle{ruled}{labelfont=normalfont,labelsep=colon,strut=off} 
\floatstyle{ruled}
\newfloat{listing}{tb}{lst}{}
\floatname{listing}{Listing}
\usepackage{times}
\usepackage{soul}
\usepackage{url}
\usepackage[utf8]{inputenc}
\usepackage{graphicx}
\usepackage{amsmath}
\usepackage{amsthm}
\usepackage{booktabs}
\usepackage[switch]{lineno}

\usepackage[table]{xcolor}
\usepackage{multirow}
\usepackage[hang,flushmargin]{footmisc}
\usepackage{textcomp} 
\usepackage[table]{xcolor}
\usepackage{graphicx}
\usepackage{epstopdf}
\usepackage{rotating}
\usepackage{newfloat}
\usepackage{listings}
\usepackage{hhline}
\usepackage{bibentry}
\usepackage{amsmath}
\usepackage{amsthm}

\usepackage{amssymb}

\usepackage{dsfont}
\usepackage{mathrsfs}

\usepackage{mathabx}
\usepackage{booktabs}

\usepackage{enumerate}

\newtheorem{definition}{Definition}
\newtheorem{thm}{Theorem}

\usepackage{amsfonts}
\usepackage{array}
\usepackage[caption=false,font=normalsize,labelfont=sf,textfont=sf]{subfig}
\usepackage{subcaption}
\usepackage{textcomp}
\usepackage{url}
\usepackage{verbatim}
\usepackage{pifont}

\title{SCoNE: Spherical Consistent Neighborhoods Ensemble for Effective and Efficient Multi-View Anomaly Detection}
\author{
    Yang Xu\textsuperscript{\rm 1,2}\thanks{These authors contributed equally.}, 
    Hang Zhang\textsuperscript{\rm 1,2}\footnotemark[1], 
    Yixiao Ma\textsuperscript{\rm 1,2}, 
    Ye Zhu\textsuperscript{\rm 3}, 
    Kai Ming Ting\textsuperscript{\rm 1,2}\thanks{Corresponding author.}
}
\affiliations{
    \textsuperscript{\rm 1} State Key Laboratory for Novel Software Technology, Nanjing University, Nanjing, China\\
    \textsuperscript{\rm 2} School of Artificial Intelligence, Nanjing University, Nanjing,  China\\
    \textsuperscript{\rm 3} School of Information Technology, Deakin University, Geelong, Australia\\
    \{xuyang, zhanghang, mayx\}@lamda.nju.edu.cn, ye.zhu@ieee.org, tingkm@nju.edu.cn
}

\begin{document}

\maketitle

\begin{abstract}
    The core problem in multi-view anomaly detection is to represent local neighborhoods of normal instances consistently across all views.
    Recent approaches consider a representation of local neighborhood in each view independently, and then capture the consistent neighbors across all views via a learning process. They suffer from two key issues. First, there is no guarantee that they can capture consistent neighbors well, especially when the same neighbors are in regions of varied densities in different views, resulting in inferior detection accuracy. Second, the learning process has a high computational cost of $\mathcal{O}(N^2)$, rendering them inapplicable for large datasets.
    To address these issues, we propose a novel method termed \textbf{S}pherical \textbf{C}onsistent \textbf{N}eighborhoods \textbf{E}nsemble (SCoNE). 
    It has two unique features: (a) the consistent neighborhoods are represented with multi-view instances directly, requiring no intermediate representations as used in existing approaches; and 
    (b) the neighborhoods have data-dependent properties, which lead to large neighborhoods in sparse regions and small neighborhoods in dense regions. The data-dependent properties enable local neighborhoods in different views to be represented well as consistent neighborhoods, without learning. This leads to $\mathcal{O}(N)$ time complexity. Empirical evaluations show that SCoNE has superior detection accuracy and runs orders-of-magnitude faster in large datasets than existing approaches. 
\end{abstract}


\section{Introduction}

Anomaly detection, \textit{a.k.a.} outlier detection, is a crucial data analysis technique~\cite{aggarwal2017introduction} with applications in diverse domains, including fraudulent transaction detection~\cite{fraudtrans09}, web spam detection~\cite{webspam11}, and network intrusion detection~\cite{netattack12}. While numerous anomaly detectors have been developed~\cite{breunig-et-al:lof,liu-ting-zhou:isolation_forest,zenati2018adversarially,coleman2019race}, these traditional detectors identify anomalies in a single-view dataset only and cannot discover multi-view anomalies, which exhibit inconsistent behaviors across different views. For example, in social media user analysis, a user can be characterized by personal attributes such as age and hobbies (view 1), and their connectivity with other users (view 2). If a user is assigned to the sports enthusiast group in one view based on their self-reported interests, but their interaction patterns in another view show engagement with non-sports related content and communities, then it is natural to consider this user's behavior anomalous~\cite{yu2015glad}. Other examples of multi-view anomalies can also be found in movie recommendation~\cite{mvad_gao11}, web image analysis~\cite{xu2013survey}, and digit recognition~\cite{li2015multi}. Thus, the goal of multi-view anomaly detection is to leverage multiple views to effectively and efficiently identify anomalies that manifest inconsistent behaviors across these views.

\begin{figure}[]
    \centering
    \includegraphics[width=0.48\textwidth]{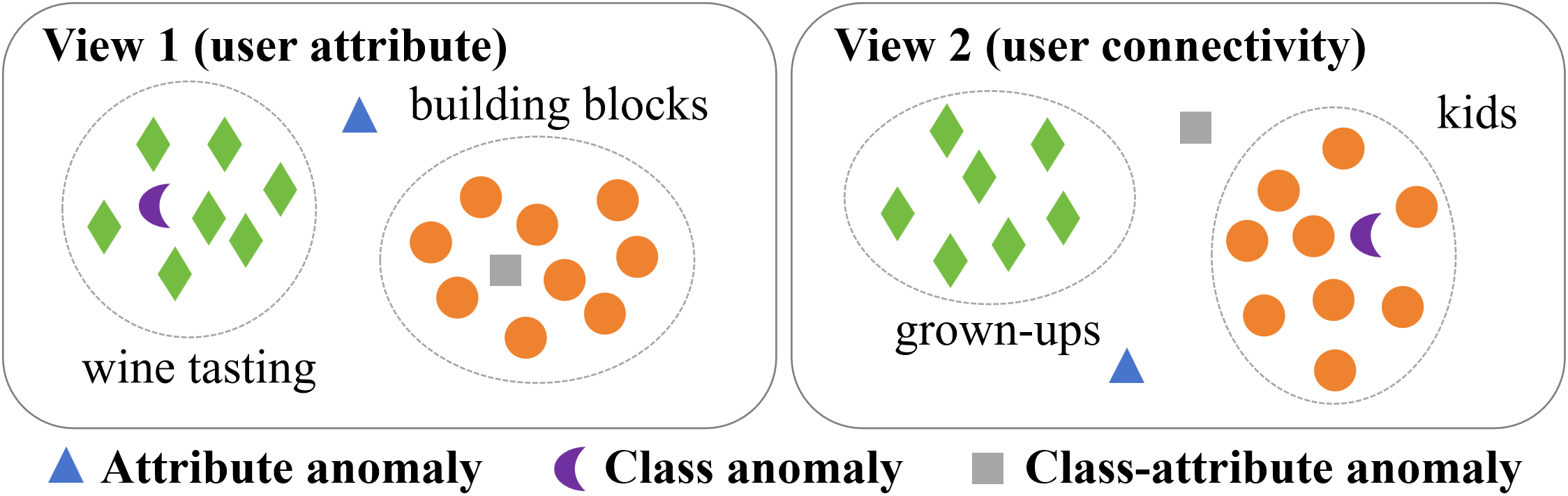}
    \caption{An illustration of three types of anomalies. 
    }
    \label{fig:anomalies1}
\end{figure} 

In single-view anomaly detection, a fundamental assumption is that normal instances have similar characteristics and constitute the majority of instances~\cite{hawkins1980identification,aggarwal2017introduction}. However, in multi-view anomaly detection, an instance which is part of the majority in every view is an anomaly if its behaviors are inconsistent across all views. Therefore, the definition of anomalies in multi-view anomaly detection is different from that in single-view anomaly detection. As introduced in the literature~\cite{mvad_ji19,HuWDZ24}, there are three different types of multi-view anomalies, as illustrated in Figure~\ref{fig:anomalies1}:
\begin{itemize}
   \item \textbf{Attribute anomaly} has different attribute characteristics from those of most other instances in each view. It is a single-view anomaly.
   \item \textbf{Class anomaly} 
   has inconsistent neighborhoods in different views, \textit{i.e.}, it may belong to different clusters or classes in different views.
   \item \textbf{Class-attribute anomaly} is the mixture of the above two types of anomalies, \textit{i.e.}, it  is an attribute anomaly in some views, but a class anomaly in other views. 
\end{itemize}

\begin{table}[]
\centering
\scalebox{0.78}{
\begin{tabular}{l|l|cccc}
\toprule
&  \textbf{Method}   & \textbf{C1} & \textbf{C2} & \textbf{C3} & \textbf{C4} \\ \midrule
\multirow{5}{*}{\rotatebox{90}{\textbf{Others}}}
& HOAD~\cite{mvad_gao11}   & \ding{55}  & \ding{55}  & \ding{55} & \ding{55}  \\
& CC~\cite{mvad_liu12}     & \ding{55}  & \ding{55}  & \ding{55} & \ding{55}  \\
& APOD~\cite{mvad_mar13}   & \ding{55}  & \ding{55}  & \ding{55} & \ding{55}  \\
& MuvAD~\cite{mvad_sheng19}  & \ding{51}  & \ding{55}  & \ding{51} & \ding{55}  \\
& HBM~\cite{mvad_wang20}    & \ding{51}  & \ding{55}  & \ding{51} & \ding{51}  \\ \midrule
\multirow{12}{*}{\rotatebox{90}{\textbf{Representation-based}}}
& DMOD~\cite{mvad_zhao15}   & \ding{51}  & \ding{51}  & \ding{55} & \ding{55}  \\
& CRMOD~\cite{mvad_zhao17}  & \ding{51}  & \ding{51}  & \ding{55} & \ding{51}  \\
& LDSR~\cite{mvad_li18}   & \ding{51}  & \ding{55}  & \ding{55} & \ding{51}  \\
& MODDIS~\cite{mvad_ji19} & \ding{51}  & \ding{55}  & \ding{51} & \ding{51}  \\
& NCMOD~\cite{mvad_cheng21}  & \ding{51}  & \ding{55}  & \ding{51} & \ding{51}  \\
& SPLSP~\cite{WangCLFZZ23}  & \ding{51}  & \ding{55}  & \ding{51} & \ding{51}  \\
& ECMOD~\cite{ChenWWHD23}  & \ding{51}  & \ding{55}  & \ding{51} & \ding{51}  \\
& MODGD~\cite{HuWDZ24}  & \ding{51}  & \ding{55}  & \ding{51} & \ding{51}  \\
& IAMOD~\cite{mvad_tkdd24}  & \ding{51}  & \ding{55}  & \ding{51} & \ding{51}  \\
& MODGF~\cite{hu2025unsupervised}  & \ding{51}  & \ding{55}  & \ding{51} & \ding{51}  \\
& SCoNE (The proposed method)   & \ding{51}  & \ding{51}  & \ding{51} & \ding{51} \\ \bottomrule
\end{tabular}}
\caption{Comparison of representative methods in terms of four capabilities ({C1}, {C2}, {C3} \& {C4}). We use ``\ding{51}" to denote that a method has a capability, and ``\ding{55}" cannot.}
\label{tab:compare1}
\end{table}

The foundational research in multi-view anomaly detection is established by pioneering methods focusing on graph and clustering inconsistencies. HOAD~\cite{mvad_gao11} constructs a combined similarity graph across all views and employs the principal eigenvectors of the graph's Laplacian matrix to identify class anomalies based on inter-component distances. Others, including CC~\cite{mvad_liu12} and APOD~\cite{mvad_mar13}, focus on detecting inconsistencies in clustering structures across views. However, a common limitation of these methods is their constraint to two-view datasets and the detection of only class anomalies.

The limitations of these early methods, coupled with the growing complexity and scale of modern data, highlight four key capabilities for a robust multi-view anomaly detection method. In the literature, these can be summarized as the capability of identifying all three types of anomalies ({C1}); running on datasets with linear time complexity ({C2}); dealing with datasets that do not have clear clustering structures ({C3}); and handling an arbitrary number of views ({C4}).
To overcome these challenges, subsequent research has made significant strides on several fronts. For example, DMOD~\cite{mvad_zhao15} uses latent coefficients and sample-specific errors to identify all types of multi-view anomalies ({C1}) with linear time complexity ({C2}), while MODDIS~\cite{mvad_ji19} employs neural networks to learn representations in a unified latent space to handle data without clear cluster structures ({C3}) and arbitrary number of views ({C4}). 
We compare existing methods and the proposed method in terms of these capabilities in Table~\ref{tab:compare1}. As no existing methods have all of these capabilities simultaneously, we are motivated to explore a method that has all capabilities.

Among these efforts, representation learning has become the most popular strategy in the field of multi-view anomaly detection because it can effectively integrate the information of different views into a consensus representation space where normal instances and anomalies can be effectively distinguished~\cite{HuWDZ24,mvad_tkdd24,hu2025unsupervised}. As the local neighborhoods of normal instances are consistent across different views, while the local neighborhoods of abnormal instances are significantly different across different views~\cite{mvad_sheng19}, the core problem of representation learning methods is to represent local neighborhoods of normal instances consistently across all views.
However, we notice that existing methods suffer from two key issues due to their shared methodology: using a representation of local neighborhood in each view independently, and then capturing the consistent neighbors across all views via a learning process. First, there is no guarantee that consistent neighborhoods are captured correctly, especially when the same neighbors are in regions of varied densities in different views. The common practice of linearly combining representations from different views disregards the complex distributions of anomalies~\cite{HuWDZ24}. This has led to inferior anomaly detection accuracy. Second, they usually has a high computational cost of $\mathcal{O}(N^2)$, preventing them from handling large-scale datasets.

Our insight is that multi-view instances can be used directly to represent consistent neighborhoods effectively across all views, without explicit learning.
We thus introduce a novel and efficient method called \textbf{S}pherical \textbf{C}onsistent \textbf{N}eighborhoods \textbf{E}nsemble (SCoNE) which incorporates this insight. 
Our major contributions are:

\begin{itemize}
    \item Proposing SCoNE which employs a small sampled set of multi-view instances to represent a set of adaptive-radius spherical regions across all views that yields consistent neighborhoods with linear time complexity. 
   \item  Revealing that SCoNE creates neighborhoods which adapt to varied local densities by leveraging two key data-dependent properties, necessary for ensuring consistent neighborhoods across all views.
    \item Conducting extensive experiments to evaluate SCoNE on both synthetic and real-world multi-view datasets. The experimental results demonstrate both the effectiveness and efficiency of SCoNE.
\end{itemize}

\section{Key Idea and Implementation}
\label{sec:met}


\textbf{Preliminaries.} Given a multi-view dataset $\mathcal{D}=\{\mathbf{x}_i|i=1,\dots,N\}$, and $\mathbf{x}_i=(\mathbf{x}_i^1,\dots,\mathbf{x}_i^V)$ is the $i$-th instance with $V$ views, where $\mathbf{x}_i^v$ denotes the $v$-th view of $\mathbf{x}_i$, and $\mathcal{D}^v=[\mathbf{x}_1^v,\dots,\mathbf{x}_N^v]$ is sample set observed in the $v$-th view.

\subsection{Key Idea: Multi-view instances are good representatives of Consistent Neighborhoods across all views} 

The concept of ``consistent neighbors" is pivotal in multi-view anomaly detection as it identifies normal instances that exhibit similar behavior across different views. 
\begin{definition} \textbf{A Consistent Neighborhood of $k$-Consistent Neighbors} 
of any $\mathbf{x} \in \mathcal{D}$ across all views is defined as:
\begin{equation}
\mathcal{CN}_k(\mathbf{x}) = \{ \mathbf{y} \in \mathcal{D} \mid \forall v \in \{1, 2, \ldots, V\}, \mathbf{y}^v \in \mathcal{N}_{k}(\mathbf{x}^v|\mathcal{D}) \}
\end{equation}
where $\mathcal{N}_k(\mathbf{x}^v|\mathcal{D}^v)$ represents the set of $k$-nearest neighbors of $\mathbf{x}^v$ in $\mathcal{D}^v$.
\label{def:CN}
\end{definition}

The above definition has two key ingredients, i.e.,
the \emph{individual-view neighborhoods} of $\mathbf{x}$ which adapt to varied local densities; and the consistent neighborhood, of  $\mathbf{x}$ being normal, is represented by the same $k$ \emph{normal multi-view instances}. A multi-view instance which violates this consistency signifies that it is likely to be a multi-view anomaly. 

Although Definition \ref{def:CN} is compatible with a common assumption in existing methods~\cite{mvad_sheng19,mvad_cheng21,WangCLFZZ23,HuWDZ24}, i.e., normal instances across different views exhibit similar local neighborhood structures, they have used representations other than the multi-view instances and with learning.
For example, to achieve this consistency, many existing approaches have employed a method  that considers a representation of local neighborhood in each view \emph{independently}, and then attempts to capture the consistent neighbors across all views via a learning process from the representations of individual views. This has two shortcomings. First, there is no guarantee that the learning can capture the consistent neighbors of a normal instance $\mathbf{x}$, especially when $\mathbf{x}$'s neighbors are in regions of varied densities in each view and the distributions vary over different views. Second, the learning process usually has high computational cost of $\mathcal{O}(N^2)$.

Instead of this indirect method, we propose a direct method and it addresses these shortcomings. 
It is direct because the consistency in the associated neighborhoods of individual views, even in varied densities, can be ensured by using a set of samples from the given multi-view dataset $\mathcal{D}$.

Specifically, with a set of representative samples $\mathcal{S}=\{\mathbf{s}_1,\dots,\mathbf{s}_{\psi}\}$ from $\mathcal{D}$,  each point $\mathbf{s}_i^v\in \mathcal{S}^v$ in view $v$ defines a neighborhood or a region containing its local normal instances.
The key is to ensure that each neighborhood represents a normal region in every view such that all these neighborhoods are represented by almost the same multi-view instances $\mathbf{s}_i \in \mathcal{S}$ across all views. This is crucial for an unbiased and right-for-the-task multi-view representation.
This method also reduces the time complexity to $\mathcal{O}(N)$ as computing the neighborhoods represented by a set of sampled instances and the final score for each $\mathbf{x}$ from the representation is straightforward, as we show in the next subsection.

\subsection{SCoNE: Multi-View  Spherical Consistent Neighborhoods Ensemble}

We introduce SCoNE, a method to determine the consistent neighborhood of $\mathbf{x}$ based on a set of adaptive-radius spherical regions, represented by a set of multi-view instances. It offers one unique advantage, i.e., the proposed adaptive spherical regions, which is a unique among existing methods, create large neighborhoods in sparse regions and small neighborhoods in dense regions in every view, while all these neighborhoods across all views are represented by the set of multi-view instances. This is essential in identifying the  consistent neighborhoods of $\mathbf{x}$ across all views. 
SCoNE has two key functions: $f$ is the single-view spherical neighborhood function in any one view; and $\mathcal{F}$ is its multi-view extension for representing local neighborhoods across all views to create a consistent neighborhood.

Given a multi-view dataset $\mathcal{D}$ of $V$ views, and let $\mathcal{S} = \{\mathbf{s}_1, \ldots, \mathbf{s}_\psi\} \subset \mathcal{D}$ be a set of $\psi$ instances, randomly drawn from $\mathcal{D}$. For each $\mathbf{s}^v_i$ in view $v$, the adaptive-radius $r^v_i$ of its spherical neighborhood is determined as:
\begin{equation}
    r^v_i = \min_{\substack{j=1,\dots,\psi \\ j \neq i}} \|\mathbf{s}^v_i - \mathbf{s}^v_j\|
\end{equation}
where $\|\mathbf{s}^v_i - \mathbf{s}^v_j\|$ denotes Euclidean distance of $\mathbf{s}^v_i$ and $\mathbf{s}^v_j$. 

\begin{definition}
     The \textbf{view-$v$
     spherical neighborhood function $f$} of any $\mathbf{x}^v\in\mathcal{D}^v$, due to $\mathbf{s}^v_i\in\mathcal{S}^v$, is defined as:
\begin{equation}
    f(\mathbf{x}^v;\mathbf{s}^v_i) = \begin{cases} 
1 & \text{if } \|\mathbf{x}^v - \mathbf{s}^v_i\| \leq r^v_i \text{ and } \mathbf{s}^v_i \in \mathcal{N}_k(\mathbf{x}^v|\mathcal{S}^v), \\
0 & \text{otherwise}. 
\label{eq:f}
\end{cases}
\end{equation}
\label{def:ksn}
\end{definition}

\begin{figure}[t!]
    \centering
    \includegraphics[width=0.44\textwidth]{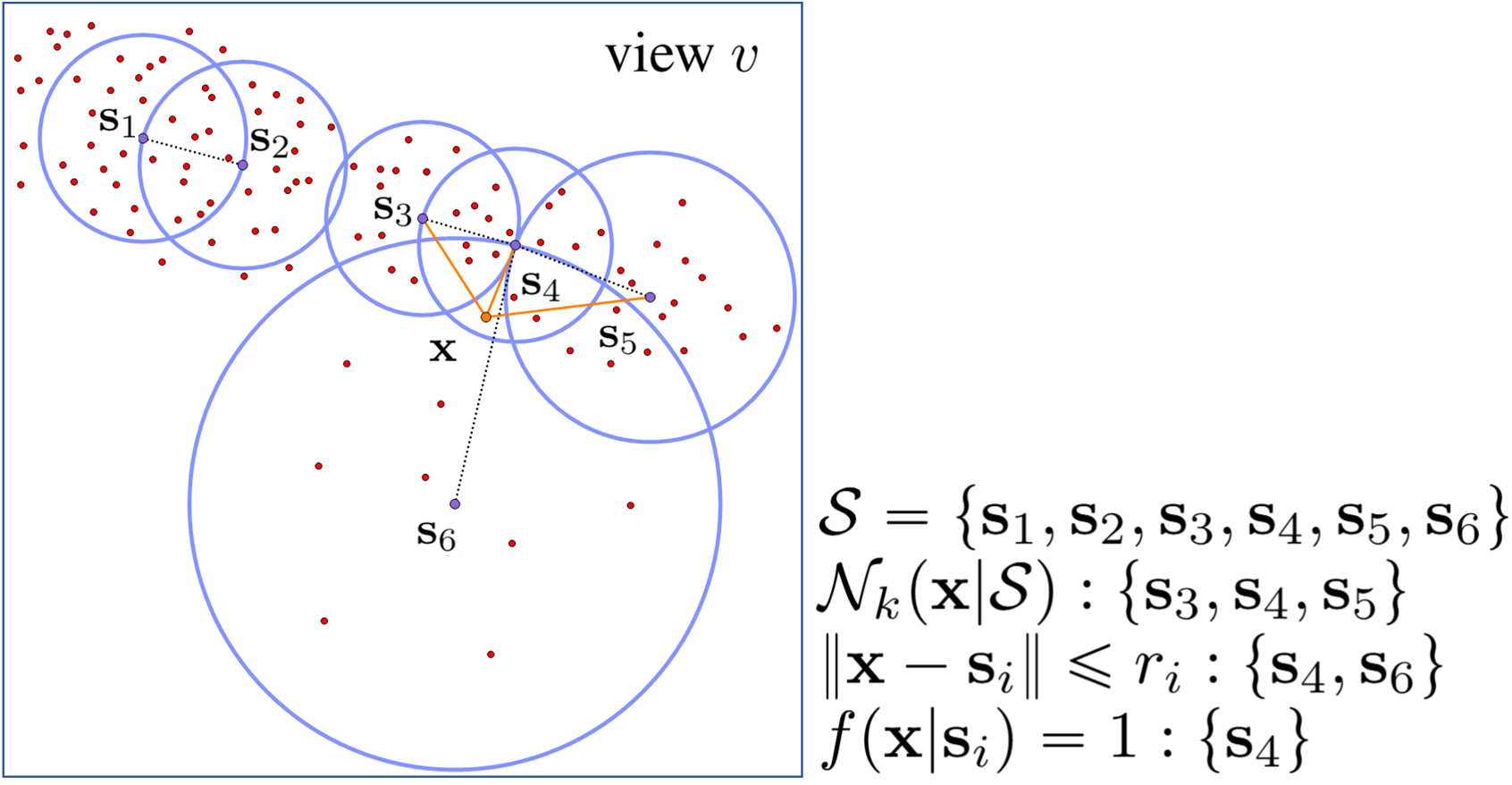}
    \caption{An illustration of the single-view spherical neighborhoods. The parameters $\psi=6$ and $k=3$ are used here. 
    }
    \label{fig:knearest4}
\end{figure} 

Each spherical neighborhood distinguishes normal instances from anomalies (being inside or outside of the neighborhood) for each $\mathbf{x}$ in each individual view, i.e., $f$ defines the membership of a single-view instance $\mathbf{x}^v$ in each spherical neighborhood, given $\mathcal{S}^v$. 
Figure~\ref{fig:knearest4} illustrates the single-view spherical neighborhoods of $k=3$ nearest neighbors of $\mathbf{x}$ derived from a set $\mathcal{S}$ of $\psi=6$ multi-view instances in view $v$.
The $k$-nearest neighbors ($k$NN) of $\mathbf{x}$ ensure that these same $k$NN of multi-view instances in $\mathcal{S}$ (which define the $k$ spherical neighborhoods) are being considered in all individual-view spherical neighborhoods in order to determine the consistency of $\mathbf{x}$ being normal across all views.

\begin{definition}
     \textbf{Multi-view  
      consistent neighborhood function $\mathcal{F}$} of any $\mathbf{x}\in\mathcal{D}$, due to $\mathbf{s}_i\in\mathcal{S}$, is defined as:
\begin{equation}
    \mathcal{F}(\mathbf{x};\mathbf{s}_i) = \prod_{v=1}^{V} f(\mathbf{x}^v;\mathbf{s}^v_i).
\label{eq:F}
\end{equation}
\label{def:multik}
\end{definition}

Note that $\mathcal{F}$ is a realization of the multi-view consistent neighborhood of $k$-consistent neighbors, stated in Definition~\ref{def:CN}. 
Specifically, $\mathcal{F}(\mathbf{x};\mathbf{s}_i) = 1$ if and only if $\mathbf{x}$ falls within the spherical neighborhood of $\mathbf{s}_i$ in all views. This product operation effectively assesses  the consistency of $\mathbf{x}$ being normal across all views: if $\mathbf{s}_i$ is not a neighbor of $\mathbf{x}$ in even one view, the product becomes $0$, indicating $\mathbf{x}$ does not have multi-view consistency of being normal. Then, by aggregating the outcomes of $t$ sets of $\mathcal{F}$ functions, SCoNE computes the consistent neighborhoods score for each instance.

\begin{definition} 
 Given $t$ sets of multi-view consistent neighborhoods generated from $t$ subsamples $\hat{\mathbf{S}} = \{\mathcal{S}_1,\dots,\mathcal{S}_t\}$,
 \textbf{the consistent neighborhoods score} of SCoNE for any $\mathbf{x}\in\mathcal{D}$, due to all $\mathcal{S}_j\in\hat{\mathbf{S}}$ of $|\mathcal{S}_j|=\psi$ points, 
 is defined as follows:
 \begin{equation}
     \bar{\mathbf{C}}(\mathbf{x}) = \frac{1}{\psi t} \sum_{j=1}^{t}\sum_{i=1}^\psi \mathcal{F}(\mathbf{x};\mathbf{s}_i|\mathcal{S}_j).
 \end{equation}
\end{definition}

The consistent neighborhood of each $\mathcal{F}$ ensures that a normal instance yields many values of $\mathcal{F}$ equal to $1$, which lead to a high $\bar{\mathbf{C}}(\mathbf{x})$ score. On the other hand, an anomaly of any type has at least some values of $\mathcal{F}$ equal to $0$, resulting in a low $\bar{\mathbf{C}}(\mathbf{x})$ score $(\ll 1)$.

For every instance in every view, the time cost $\mathcal{O}(\psi t k)$ is needed to compute $f$. Thus, SCoNE has the overall time complexity $\mathcal{O}(\psi tkVN)$ which is linear to the dataset size $N$.
The effectiveness of the consistent neighborhoods score derives directly from the observation that normal instances and anomalies are distinguishable in a representation space (through a conceptual understanding of the consistent neighborhood) generated by $\mathcal{F}$, as detailed in the next subsection.


\subsection{Conceptual understanding of Consistent Neighborhood}

The following offers a conceptual understanding of Consistent Neighborhood, stated in Definition \ref{def:CN}, via function $f$.

\begin{figure}[]
  \centering
  \begin{minipage}{0.48\textwidth}
    \includegraphics[width=\linewidth]{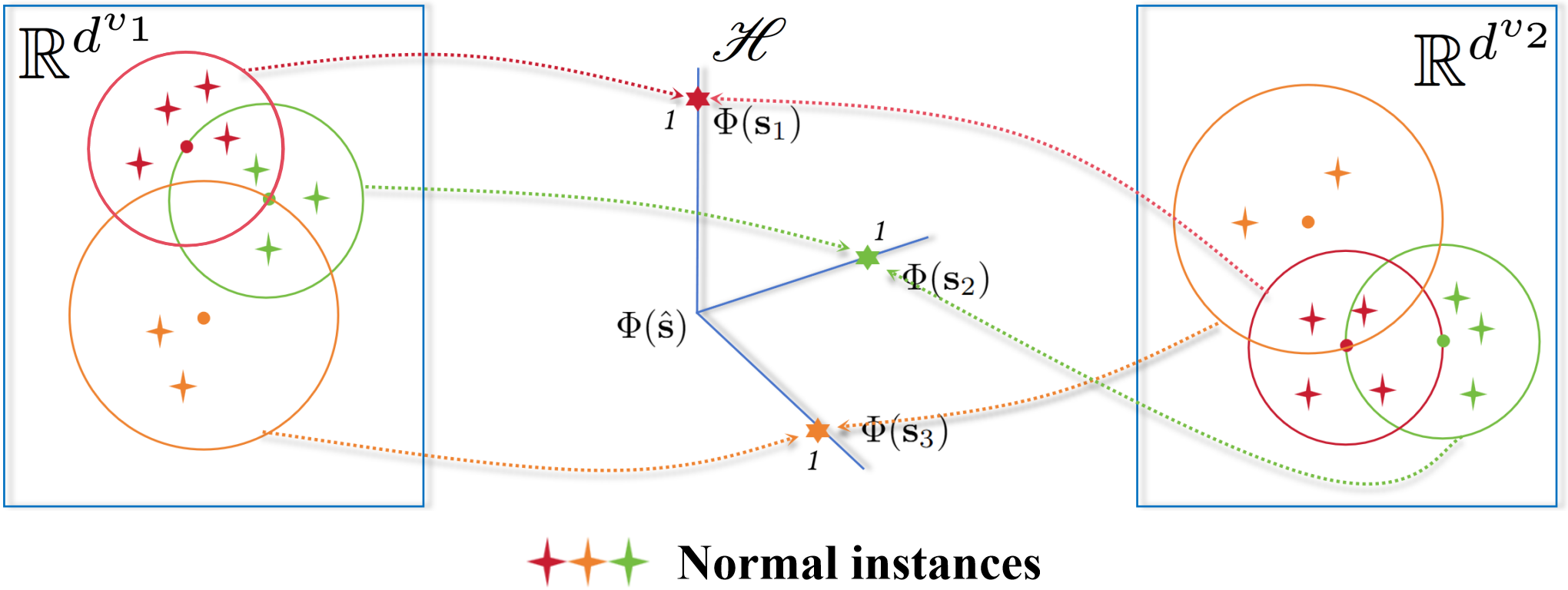}
    \caption{
    An illustration of the mapping $\Phi$ for normal instances in $\mathscr{H}$. Three sampled points $\mathbf{s}_1,\mathbf{s}_2,\mathbf{s}_3\in \mathcal{S}$ (represented as dots), define three spherical neighborhoods of normal regions in each view.
    }
    \label{fig:image1}
  \end{minipage}
  \hfill
  \vspace*{3mm}
  \begin{minipage}{0.48\textwidth}
    \includegraphics[width=\linewidth]{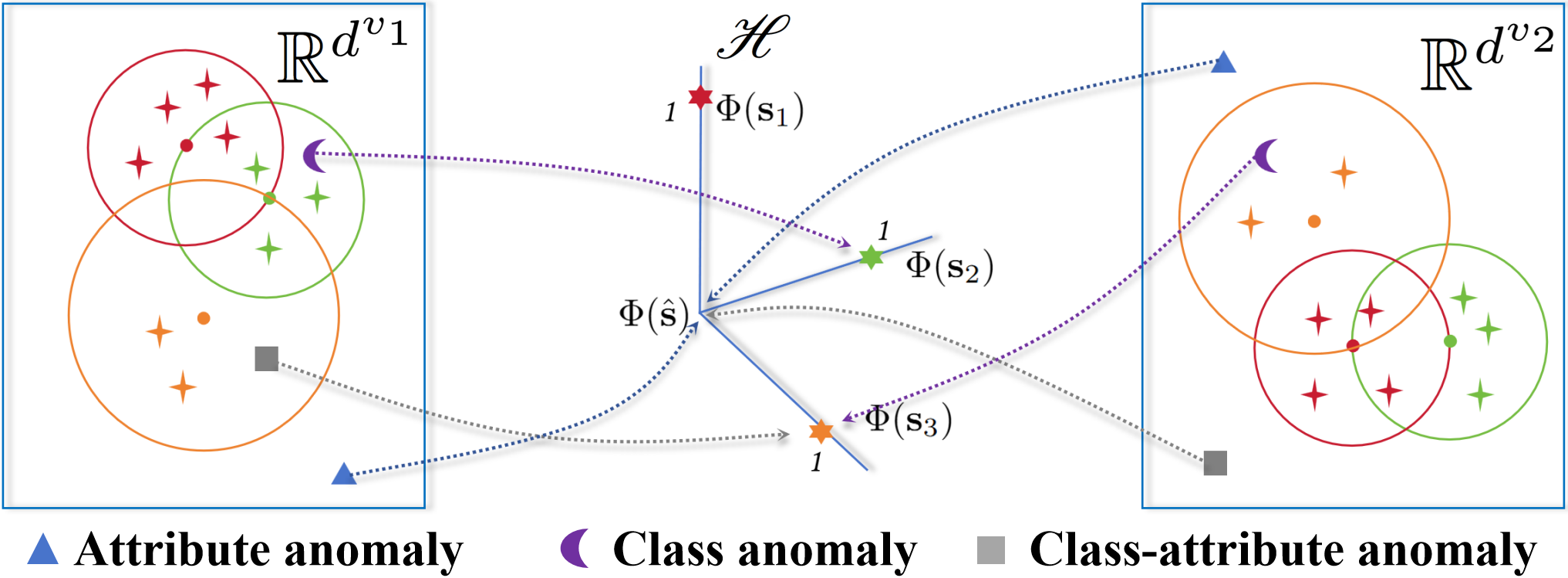}
    \caption{An illustration of the mapping $\Phi$ for three types of multi-view anomalies. Each violates a different kind of consistency of being normal.}
    \label{fig:image2}
  \end{minipage}
\end{figure}

Let $\Phi :\mathbf{x}^v \mapsto \{0,1\}^{\psi}$ be the mapping of $f$ that transforms an instance $\mathbf{x}^v$ into a $\psi$-dimensional binary space $\mathscr{H}$. Specifically, $\Phi(\mathbf{x}^v)=[f(\mathbf{x}^v;\mathbf{s}^v_1),\dots,f(\mathbf{x}^v;\mathbf{s}^v_\psi)]$, where dimension $i$ indicates membership in the corresponding spherical neighborhood of $\mathbf{s}^v_i$. Let $\Phi({\mathbf{s}}_i)$ and $\Phi(\hat{\mathbf{s}})$ denote the representations of $\mathbf{s}_i$ and the origin in the representation space $\mathscr{H}$, respectively. Figure~\ref{fig:image1} and Figure~\ref{fig:image2} show an example of the mapping $\Phi$ with parameters $V=2,t=1,k=1$ and $\psi=3$. In $\mathscr{H}$, normal instances and anomalies are distinguishable based on the following interpretations.

Normal instances, due to their consistent behavior across all views, are likely to be nearest neighbors of the same sampled instance $\mathbf{s}_i$ in all views. Consequently, they are mapped to the same position $\Phi({\mathbf{s}}_i)$ in $\mathscr{H}$. In contrast, anomalies exhibit inconsistent  neighborhoods across different views. Specifically: (1) Attribute anomalies are unlikely to be nearest neighbors of any $\mathbf{s}\in\mathcal{S}$ in any view; thus they are mapped to the origin $\Phi(\hat{\mathbf{s}})$. (2) A class anomaly is a nearest neighbor to different samples in $\mathcal{S}$ in different views (e.g., $\mathbf{s}_i$ in one view and $\mathbf{s}_j$ in another, where $i\neq j$), resulting in mappings to distinct positions in $\mathscr{H}$. (3) Class-attribute anomalies exhibit characteristics of attribute anomaly in some views, and those of class anomaly in others.

\subsection{Data-dependent Properties: Capturing Consistent Neighborhoods in Varied Densities}

To guarantee the capture of consistent neighborhoods across all views, the spherical neighborhoods generated by SCoNE have the following two key properties.

\begin{thm}
Given a multi-view dataset $\mathcal{D}$, and let $\mathcal{S}\subset\mathcal{D}$ be a set of $\psi$ sampled points. For each $\mathbf{s}^v\in\mathcal{S}^v$, let $\theta(\mathbf{s}^v)$ denote the neighborhood generated by $\mathbf{s}^v$, and $\mathcal{R}(\mathbf{s}^v)$ denote the set of all normal instances from $\mathcal{D}$ that within $\theta(\mathbf{s}^v)$ in view $v$. Then, for any two views $v_1$ and $v_2$ of $\mathcal{D}$, $\mathcal{R}(\mathbf{s}^{v_1})$ and $\mathcal{R}(\mathbf{s}^{v_2})$ contain essentially the same instances, i.e.,
\begin{equation}
    \mathbb{E}[|\mathcal{R}(\mathbf{s}^{v_1})|]\simeq\mathbb{E}[|\mathcal{R}(\mathbf{s}^{v_2})|]
\end{equation}
where $\mathbb{E}[|\cdot|]$ is the expected number of the set.
\label{thm:t2}
\end{thm}

Theorem~\ref{thm:t2} establishes that, for each sampled point, the spherical neighborhoods constructed across different views provide an equivalent estimation of the number of normal instances. This property is significant because it ensures that the expected consistent neighborhoods score for any normal instance is equivalent, as normal instances are equally likely to be contained within any spherical neighborhood, thus yielding unbiased estimates across all views.

\begin{thm}
    Given two datasets $\mathcal{D}$ and ${\mathcal{D}'}$ with same number of points observed in view of $v$, where each point in $\mathcal{D}$ and ${\mathcal{D}'}$ belongs to a subspace $\mathcal{X} \subseteq \mathbb{R}^{d}$ and is drawn from probability distributions $\mathcal{P}_{\mathcal{D}}$ and $\mathcal{P}_{{\mathcal{D}'}}$ defined on $\mathbb{R}^{d}$, respectively. Both $\mathcal{P}_{\mathcal{D}}$ and $\mathcal{P}_{{\mathcal{D}'}}$ are strictly positive on $\mathcal{X}$. Let $\mathcal{E} \subset \mathcal{X}$ be a region such that for all $\mathbf{x} \in \mathcal{E}$, $\mathcal{P}_{\mathcal{D}}(\mathbf{x}) < \mathcal{P}_{{\mathcal{D}'}}(\mathbf{x})$, \textit{i.e.}, $\mathcal{D}$ is sparser than ${\mathcal{D}'}$ in $\mathcal{E}$. Given two randomly sampled sets $\mathcal{S} \subset \mathcal{D}$ and ${\mathcal{S}'} \subset {\mathcal{D}'}$, where $|\mathcal{S}| = |{\mathcal{S}'}| = \psi$. 
    Assume that there exists a point $\mathbf{s} \in \{{\mathcal{S}} \cap {\mathcal{S}'}\}$.
    Then, for any $\mathbf{x}\in\mathcal{E}$, the $k$-nearest neighborhoods function $f$ have the property that 
    \begin{equation}
    P(f(\mathbf{x};\mathbf{s}_i|\mathcal{S})=1)>P(f(\mathbf{x};\mathbf{s}_i|\mathcal{S'})=1),
    \end{equation}
    where $f(\mathbf{x};\mathbf{s}_i|\mathcal{S})$ denotes $f(\mathbf{x}^v;\mathbf{s}^v_i)$ based on the radius of $\mathbf{s}_i^v$ calculated through the sample set $\mathcal{S}^v$. For simplicity, we omit the requisite $v$ in most notations.
\label{thm:t1}
\end{thm}

Theorem~\ref{thm:t1} shows that, SCoNE creates appropriately sized neighborhoods by adapting to local data densities: producing small neighborhoods in dense regions and large neighborhoods in sparse region. It is a necessary requirement because a multi-view dataset can have varied local densities in each view, and the distribution can vary significantly from one view to another. Thus, any neighborhood representation must represent the normal regions correctly so that the consistency of being normal across all views can be assessed correctly.
Existing methods have difficulty getting this right. They rely on learning a combination of independent representations of individual views, and this could not ensure that all normal regions are presented appropriately over all views.
Due to space limitations, the proofs of all theorems are provided in the Appendix.

\section{Experiment}
\label{sec:exp}

This section empirically validates the claimed effectiveness and efficiency of SCoNE through a comprehensive comparison with established state-of-the-art methods.
Our evaluation is conducted across a wide range of datasets, including synthetic datasets with varied densities, multiple benchmark datasets, and a multi-view social network dataset.

\subsection{Competing Methods and Setups}

We compare the performance of SCoNE against several baselines: iForest~\cite{liu-ting-zhou:isolation_forest}, MODDIS~\cite{mvad_ji19}, HBM~\cite{mvad_wang20}, NCMOD~\cite{mvad_cheng21}, ECMOD~\cite{ChenWWHD23}, MODGD~\cite{HuWDZ24}, IAMOD~\cite{mvad_tkdd24}, and MODGF~\cite{hu2025unsupervised}. Their capabilities and categories are detailed in Table~\ref{tab:compare1}. Among these, MODDIS and NCMOD are \textit{deep learning-based} methods, and HBM is the only \textit{semi-supervised learning} method. Notably, iForest serves as a benchmark for single-view anomaly detection applied to multi-view data, where multiple views are concatenated into a single view before input.

To ensure a fair comparison, we set the parameter configuration in the original works for comparison methods. For SCoNE, we fix the ensemble size $t$ to 200. The number of subsamples $\psi$ is searched in $ \{2^i|i=1,2,\dots,8\}$. The value of $k$ is searched in $\{1,3,5,7,11,21,51,101\}$. We adopted the Area Under the ROC Curve (AUC) as the evaluation metric, acknowledging its robustness in anomaly detection tasks with imbalanced datasets. The experiments are executed on a Linux CPU machine: AMD 128-core CPU with each core running at 2 GHz and  1T GB RAM.

\subsection{Synthetic Data}
\label{sec:syn}

\begin{figure}[]
    \centering
    \includegraphics[width=0.46\textwidth]{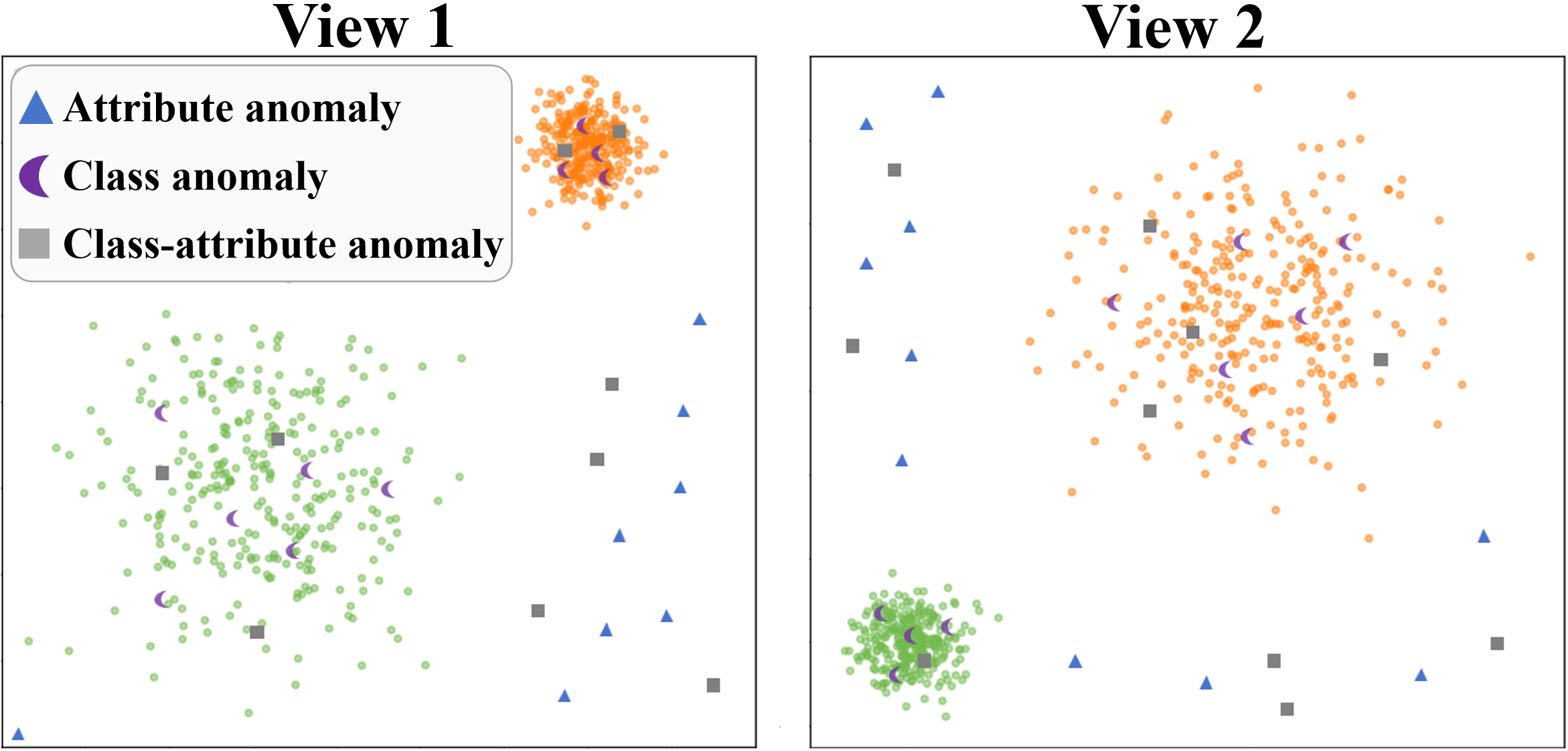}
    \caption{Synthetic datasets with various densities of distributions. Both views derive from the same original instances.
    }
    \label{fig:syndatasets}
\end{figure}

We first evaluate SCoNE using synthetic data with clusters of varied densities. Using scikit-learn, we generate a two-view dataset of $N=1000$ instances: 970 normal instances, 10 attribute anomalies, 10 class anomalies, and 10 class-attribute anomalies. This dataset generation process is repeated twenty times to ensure a robust evaluation. Figure~\ref{fig:syndatasets} illustrates a representative synthetic dataset. To isolate the effect of density variations on anomaly detection, we generate twenty control datasets with clusters of same density. 

The results presented in Table~\ref{tab:synthetic} demonstrate that SCoNE achieves the highest AUC scores on all datasets of uniform and varied densities. This is because SCoNE's spherical neighborhoods, which adapt to varied local densities in every view, enable it to perform well on both datasets. In contrast, the performance of other methods degrades on datasets with varied densities compared to those with uniform density. As expected, the single-view detector iForest has lower AUC than any of the multi-view methods.

To further investigate the reason behind SCoNE's robust performance, we evaluate the ability of representation-based methods to capture consistent neighbors. Specifically, for a normal instance $\mathbf{x}$, we compute its consistent neighbors $\mathcal{CN}_k(\mathbf{x})$ using $k=200$. Then, we calculate the proportion of these consistent neighbors that are still the $k$-nearest neighbors in the learned representation space (or the optimized similarity matrix). We randomly selected 20 normal instances and reported the average of the proportion values in Table~\ref{tab:synthetic}. The results confirm our findings, showing that SCoNE maintains the highest proportion of consistent neighbors on both datasets, which explains the reason for its robust and superior AUC performance.


\begin{table}[]
\centering
\resizebox{0.98\linewidth}{!}{
\begin{tabular}{c|c|c|c|c}
\toprule
        & \multicolumn{2}{c|}{Uniform density}     & \multicolumn{2}{c}{Varied densities}    \\ \cline{2-5}
        & AUC   & Proportion   & AUC   & Proportion \\ \midrule
iForest & $0.896_{\pm 0.010}$ & - & $0.835_{\pm 0.014}$  & -  \\ 
MODDIS  & $0.997_{\pm 0.002}$ & $85.5$ & $0.952_{\pm 0.006}$  & $78.6$ \\
HBM     & $0.981_{\pm 0.004}$ & - & $0.925_{\pm 0.008}$  & - \\
NCMOD   & $0.986_{\pm 0.004}$ & $81.7$ & $0.932_{\pm 0.010}$  & $74.9$ \\
ECMOD   & $0.977_{\pm 0.003}$ & $79.1$ & $0.933_{\pm 0.005}$  & $75.4$  \\
MODGD   & $0.954_{\pm 0.005}$ & $76.8$  & $0.907_{\pm 0.008}$  & $69.1$ \\ 
IAMOD   & $0.995_{\pm 0.005}$ & $82.3$ & $0.936_{\pm 0.007}$  & $77.9$ \\ 
MODGF   & $0.994_{\pm 0.003}$ & $83.5$ & $0.940_{\pm 0.007}$  & $76.3$ \\
\midrule
SCoNE   & \textbf{1.000}$_{\pm 0.000}$ & $\textbf{87.6}$ & \textbf{1.000}$_{\pm 0.000}$ & $\textbf{86.9}$ \\ \bottomrule
\end{tabular}}
\caption{Comparison on synthetic datasets. The mean AUC ± 2*SE (Standard Error) are shown in the table. The proportion results are presented as percentages (\%).}
\label{tab:synthetic}
\end{table}



\begin{table}[t!]
\centering
\resizebox{1\linewidth}{!}{
\begin{tabular}{lcccccc}
\toprule
           & \texttt{Zoo} & \texttt{Park.} & \texttt{Wdbc} & \texttt{MNIST}  & \texttt{AWA}              & \texttt{Caltech}              \\ \midrule
\textbf{Views}      & 3   & 3          & 3    & 3      & 6                   & 6                      \\
\textbf{Instances}  & 101 & 197        & 569  & {70,000} & 800                 & 1,474                  \\
\textbf{Clusters} & 7   & 2          & 2    & 10     & 10                  & 7                      \\
\textbf{Dimensions} & 16  & 22         & 32   & 786    & {10,940}              & 3,766                 \\ \bottomrule
\end{tabular}}
\caption{ Datasets characteristics.}
\label{tab:dataset_details}
\end{table}

\begin{table*}[]
\resizebox{1\linewidth}{!}{
\begin{tabular}{c|cccccccc|c}
\toprule
      & iForest   & MODDIS    & HBM       & NCMOD     & ECMOD     & MODGD     & IAMOD     & MODGF     & SCoNE      \\ \midrule
\texttt{Z}\_258 & $0.763_{\pm 0.027}$ & $0.818_{\pm 0.016}$ & $0.832_{\pm 0.017}$ & $0.846_{\pm 0.018}$ & $0.805_{\pm 0.024}$ & $0.816_{\pm 0.014}$ & $0.853_{\pm 0.013}$ & $0.875_{\pm 0.011}$ & \textbf{0.881}$_{\pm 0.010}$ \\
\texttt{Z}\_555 & $0.791_{\pm 0.031}$ & $0.854_{\pm 0.014}$ & $0.878_{\pm 0.015}$ & $0.881_{\pm 0.014}$ & $0.842_{\pm 0.021}$ & $0.863_{\pm 0.011}$ & $0.887_{\pm 0.011}$ & $0.899_{\pm 0.010}$ & \textbf{0.903}$_{\pm 0.010}$ \\
\texttt{Z}\_852 & $0.829_{\pm 0.025}$ & $0.861_{\pm 0.012}$ & $0.891_{\pm 0.011}$ & $0.893_{\pm 0.012}$ & $0.869_{\pm 0.011}$ & $0.905_{\pm 0.010}$ & $0.919_{\pm 0.009}$ & $0.922_{\pm 0.009}$ & \textbf{0.924}$_{\pm 0.009}$ \\
\texttt{P}\_258 & $0.717_{\pm 0.033}$ & $0.806_{\pm 0.012}$ & $0.794_{\pm 0.016}$ & $0.788_{\pm 0.014}$ & $0.781_{\pm 0.021}$ & $0.752_{\pm 0.015}$ & $0.832_{\pm 0.016}$ & $0.841_{\pm 0.014}$ & \textbf{0.849}$_{\pm 0.013}$ \\
\texttt{P}\_555 & $0.758_{\pm 0.031}$ & $0.867_{\pm 0.008}$ & $0.843_{\pm 0.013}$ & $0.863_{\pm 0.010}$ & $0.868_{\pm 0.017}$ & $0.867_{\pm 0.014}$ & $0.895_{\pm 0.014}$ & \textbf{0.918}$_{\pm 0.013}$ & ${0.912}_{\pm 0.012}$ \\
\texttt{P}\_852 & $0.822_{\pm 0.028}$ & $0.949_{\pm 0.006}$ & $0.912_{\pm 0.009}$ & $0.943_{\pm 0.006}$ & $0.937_{\pm 0.009}$ & $0.905_{\pm 0.012}$ & $0.960_{\pm 0.012}$ & $0.958_{\pm 0.010}$ & \textbf{0.962}$_{\pm 0.009}$ \\
\texttt{W}\_258 & $0.623_{\pm 0.037}$ & $0.835_{\pm 0.008}$ & $0.851_{\pm 0.009}$ & $0.839_{\pm 0.014}$ & $0.838_{\pm 0.026}$ & $0.834_{\pm 0.010}$ & $0.863_{\pm 0.012}$ & $0.867_{\pm 0.010}$ & \textbf{0.869}$_{\pm 0.009}$ \\
\texttt{W}\_555 & $0.705_{\pm 0.031}$ & $0.900_{\pm 0.006}$ & $0.907_{\pm 0.006}$ & $0.902_{\pm 0.006}$ & $0.896_{\pm 0.015}$ & $0.884_{\pm 0.007}$ & $0.910_{\pm 0.010}$ & $0.911_{\pm 0.008}$ & \textbf{0.912}$_{\pm 0.007}$ \\
\texttt{W}\_852 & $0.831_{\pm 0.026}$ & $0.955_{\pm 0.004}$ & $0.954_{\pm 0.005}$ & $0.957_{\pm 0.004}$ & $0.961_{\pm 0.017}$ & $0.946_{\pm 0.007}$ & $0.957_{\pm 0.009}$ & $0.955_{\pm 0.008}$ & \textbf{0.962}$_{\pm 0.007}$ \\
\texttt{M}\_258 & $0.614_{\pm 0.061}$ & $0.938_{\pm 0.002}$ & $0.853_{\pm 0.011}$ & $0.917_{\pm 0.006}$ & $0.887_{\pm 0.006}$ & $0.897_{\pm 0.007}$ & $0.931_{\pm 0.005}$ & $0.939_{\pm 0.004}$ & \textbf{0.943}$_{\pm 0.002}$ \\
\texttt{M}\_555 & $0.677_{\pm 0.063}$ & $0.904_{\pm 0.001}$ & $0.839_{\pm 0.003}$ & $0.893_{\pm 0.002}$ & $0.861_{\pm 0.003}$ & $0.871_{\pm 0.004}$ & $0.909_{\pm 0.001}$ & $0.913_{\pm 0.001}$ & \textbf{0.916}$_{\pm 0.001}$ \\
\texttt{M}\_852 & $0.602_{\pm 0.059}$ & $0.875_{\pm 0.009}$ & $0.762_{\pm 0.017}$ & $0.803_{\pm 0.012}$ & $0.835_{\pm 0.011}$ & $0.762_{\pm 0.014}$ & $0.856_{\pm 0.009}$ & $0.885_{\pm 0.008}$ & \textbf{0.897}$_{\pm 0.006}$ \\
\texttt{A}\_258 & $0.665_{\pm 0.042}$ & $0.783_{\pm 0.022}$ & $0.761_{\pm 0.027}$ & $0.871_{\pm 0.023}$ & $0.779_{\pm 0.031}$ & $0.851_{\pm 0.026}$ & $0.901_{\pm 0.016}$ & $0.912_{\pm 0.013}$ & \textbf{0.918}$_{\pm 0.011}$ \\
\texttt{A}\_555 & $0.713_{\pm 0.037}$ & $0.821_{\pm 0.019}$ & $0.797_{\pm 0.023}$ & $0.897_{\pm 0.021}$ & $0.845_{\pm 0.026}$ & $0.879_{\pm 0.023}$ & $0.924_{\pm 0.014}$ & $0.935_{\pm 0.011}$ & \textbf{0.941}$_{\pm 0.009}$ \\
\texttt{A}\_852 & $0.734_{\pm 0.033}$ & $0.819_{\pm 0.021}$ & $0.777_{\pm 0.024}$ & $0.883_{\pm 0.018}$ & $0.829_{\pm 0.027}$ & $0.867_{\pm 0.019}$ & $0.913_{\pm 0.013}$ & $0.928_{\pm 0.009}$ & \textbf{0.933}$_{\pm 0.006}$ \\
\texttt{C}\_258 & $0.796_{\pm 0.035}$ & $0.861_{\pm 0.029}$ & $0.786_{\pm 0.025}$ & $0.907_{\pm 0.022}$ & $0.845_{\pm 0.024}$ & $0.903_{\pm 0.021}$ & $0.943_{\pm 0.012}$ & $0.958_{\pm 0.010}$ & \textbf{0.967}$_{\pm 0.008}$ \\
\texttt{C}\_555 & $0.759_{\pm 0.039}$ & $0.836_{\pm 0.027}$ & $0.779_{\pm 0.031}$ & $0.888_{\pm 0.025}$ & $0.806_{\pm 0.029}$ & $0.878_{\pm 0.024}$ & $0.899_{\pm 0.017}$ & $0.915_{\pm 0.014}$ & \textbf{0.927}$_{\pm 0.010}$ \\
\texttt{C}\_852 & $0.782_{\pm 0.030}$ & $0.845_{\pm 0.026}$ & $0.797_{\pm 0.024}$ & $0.911_{\pm 0.020}$ & $0.832_{\pm 0.024}$ & $0.914_{\pm 0.021}$ & $0.927_{\pm 0.008}$ & $0.938_{\pm 0.008}$ & \textbf{0.945}$_{\pm 0.007}$ \\ 
\midrule
\textbf{Avg. rank} & 8.94 & 5.64 & 6.86 & 4.75 & 6.33 & 6.17 & 3.08 & 2.31  & \textbf{1.06} \\ \bottomrule
\end{tabular}}
\caption{The mean AUC ± 2*SE results on identifying three types of anomalies on benchmark datasets. For brevity in presentation, we denote each dataset using the initial letter of its name combined with the percentages of three anomaly types. For example, ``\texttt{Z}\_258" denotes the \texttt{Zoo} dataset with 2\% attribute anomalies, 5\% class-attribute anomalies and 8\% class anomalies.}
\label{tab:mainall}
\end{table*}

\subsection{Benchmark Datasets}
\label{subsec:bench}


This experiment evaluates SCoNE using multiple benchmarks from the UCI Machine Learning Repository and real-world multi-view applications. While dataset choices can vary across studies, UCI datasets.
remain widely adopted for evaluation in this field. Their use ranges from pioneering work like HOAD~\cite{mvad_gao11} to many recent methods such as~\cite{mvad_cheng21, ChenWWHD23, WangCLFZZ23, mvad_tkdd24}. 
To maintain consistency with frequently used benchmarks and cover varied data scales and dimensions, we employ four UCI datasets: \texttt{Zoo}, \texttt{Parkinson}, \texttt{Wdbc}, and \texttt{MNIST}. 
Furthermore, aligning with recent work~\cite{mvad_tkdd24}, we evaluate on two challenging real-world multi-view datasets, \texttt{AWA}
and \texttt{Caltech}.
These were chosen for their inherent high dimensionality and availability of six distinct views. 
Key characteristics of all the datasets are summarized in Table~\ref{tab:dataset_details}. 


For the UCI datasets, three views are generated by splitting the original features into three subsets, following common practice~\cite{HuWDZ24, mvad_tkdd24}. 
The \texttt{AWA} and \texttt{Caltech} datasets are used with their original six views. 
To generate anomalies, we strictly adhere to the procedures established in existing studies~\cite{ChenWWHD23,mvad_tkdd24}. Specifically, three types of anomalies are created: 
(i) \textit{Attribute anomalies}, by randomly altering features of selected samples; 
(ii) \textit{Class-attribute anomalies}, by first swapping features in a partial set of views between chosen sample pairs, followed by replacing features in the remaining views with random data; 
(iii) \textit{Class anomalies}, by randomly selecting pairs of samples and exchanging features between a partial set of their views. 
On each dataset, we repeat this procedure twenty times to ensure robust evaluation.

\begin{figure}[]
    \centering
    \includegraphics[width=0.33\textwidth]{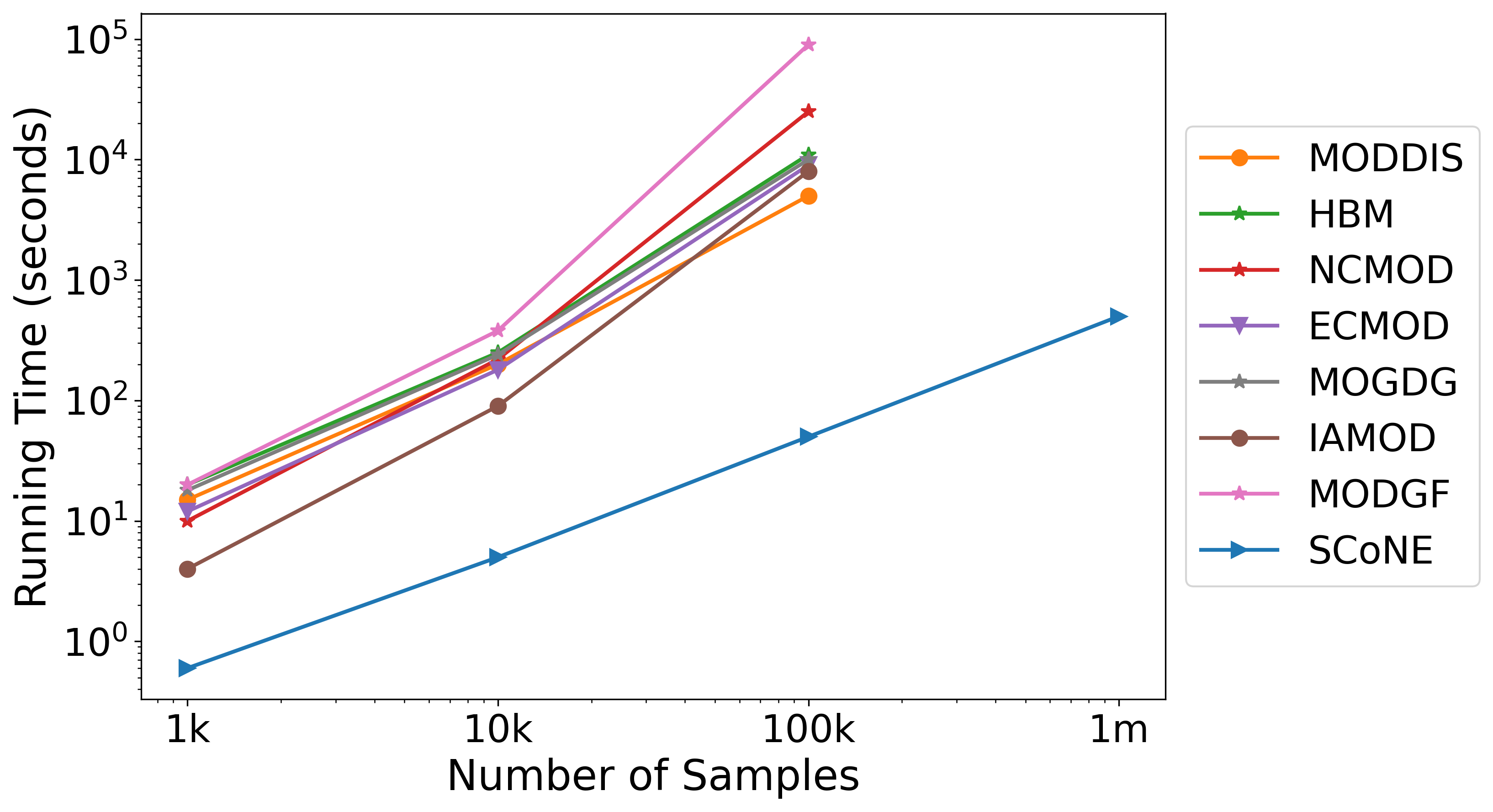}
    \caption{The runtime comparison on \texttt{MNIST} dataset.
    }
    \label{fig:runtime}
\end{figure}

The results in Table~\ref{tab:mainall} demonstrate the superior performance of SCoNE, which achieves the best results with an average rank of $1.06$. The single-view baseline, iForest, predictably performs the worst, confirming the necessity of leveraging multi-view information. It is also noteworthy that HBM, despite being a semi-supervised method that utilizes partial label information, is generally outperformed by most unsupervised representation-based methods. Among all competing methods, MODGF emerges as the closest competitor. Its strong performance can be attributed to its use of graph filtering, a technique that enhances the representation of consistent neighborhoods by effectively denoising each view. 
To further assess runtime performance on large-scale data, we utilize an expanded version of the \texttt{MNIST} dataset, which was the largest among our initially selected benchmarks. 
Specifically, starting from the standard \texttt{MNIST} dataset, we generated a significantly larger dataset by applying the data expansion methodology described in~\cite{loosli-canu-bottou-2006}. 
This allowed for a thorough evaluation of scalability. 
We constructed three views for this expanded dataset, consistent with our setup for the original \texttt{MNIST}, and assessed all methods under varying numbers of samples drawn from this large-scale version. 
As shown in Figure~\ref{fig:runtime}, SCoNE exhibits significantly faster runtime, operating at least two orders of magnitude faster than other methods across different scales. 
Notably, at the scale of one million instances, other methods failed to produce results within a 24-hour period, whereas SCoNE completed the run efficiently in 428 seconds.

\begin{figure}[]
    \centering
    \includegraphics[width=0.46\textwidth]{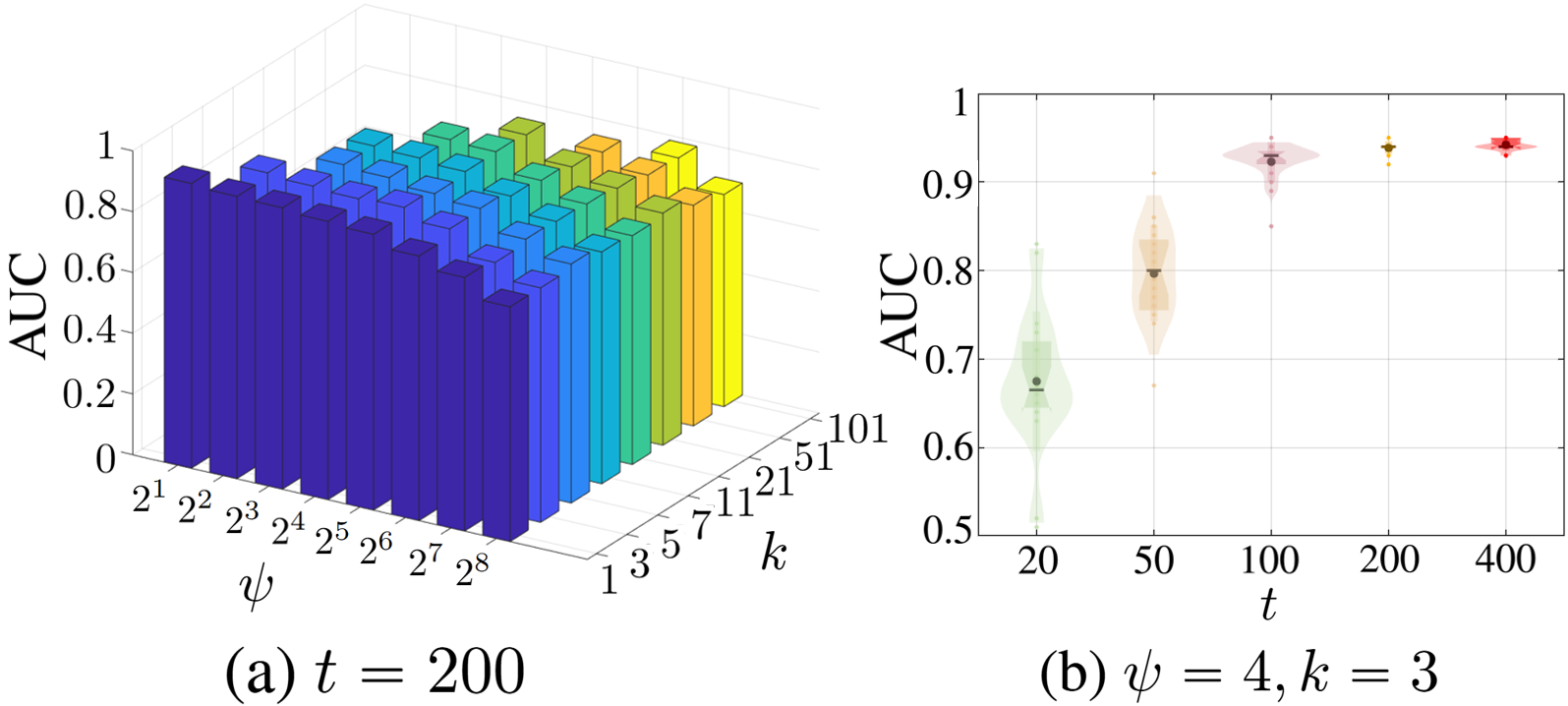}
    \caption{Analytical experiments on the \texttt{MNIST} (\texttt{M}\_258) dataset. (a) The AUC values with different values of $\psi$ and $k$. (b) The AUC values with different values of $t$.
    }
    \label{fig:paraanaly}
\end{figure}

\textbf{Parameter analysis.} The performance of  SCoNE is influenced by three parameters: $\psi,k$ and $t$. Figure~\ref{fig:paraanaly}(a) shows the results of varying $\psi$ and $k$, and Figure~\ref{fig:paraanaly}(b) shows the impact of changing $t$. Our analysis reveals that while $\psi$ and $k$ can be easily adjusted within a reasonable range,  tuning is necessary for peak performance. Notably, the ensemble parameter $t >100$  yielded satisfactory results, with increasing $t$ leading to decreased variance and robust results.

\subsection{Ablation Study}

We conduct an ablation study to evaluate the contribution of \textit{\textbf{spherical neighborhoods}} and \textit{\textbf{$k$-nearest neighbors}} to the effectiveness of SCoNE. Specifically, we conduct empirical experiments to assess the performance of \textit{\textbf{SCoNE-VD}}, which creates neighborhoods via Voronoi diagram~\cite{aurenhammer1991voronoi}, a nearest-neighbor-based partitioning mechanism, and \textit{\textbf{SCoNE-1NN}}, which employs only the one nearest neighbor, in identifying different types of anomalies. The comparison results are shown in Table~\ref{tab:ablation}.
The results demonstrates that both spherical neighborhoods and $k$-nearest neighbors clearly contribute to the performance of SCoNE, which is due to the fact that spherical neighborhood distinguishes normal instances from anomalies and $k$-nearest neighbors determine the consistency across all views.

\subsection{Anomaly Detection in Social Networks}

To evaluate the performance of SCoNE for anomaly detection in social networks, we conduct experiment on the \texttt{Facebook}
dataset~\cite{mcauley-leskovec:learning_discover_social_circles}, which is commonly used for social network analysis tasks. \texttt{Facebook} contains 4,039 users and 88,234 undirected edges, which together form ten distinct social networks. In social network anomaly detection, a user is considered normal if they consistently belong to the same social network across different views of the data. Conversely, a user is classified as a class anomaly if their social network membership varies across views~\cite{yu2015glad}. To establish a baseline for comparison, we employ LDSCEN, a method designed for social network community detection~\cite{mcauley-leskovec:learning_discover_social_circles}. For LDSCEN, if an instance is classified as the same social network across all views, it is identified to be normal; otherwise, it is identified as an anomaly.

\begin{table}[]
\centering
\resizebox{0.98\linewidth}{!}{
\begin{tabular}{l|cccccc}
\toprule
          & \texttt{A}\_258 & \texttt{A}\_555 & \texttt{A}\_852 & \texttt{C}\_258 & \texttt{C}\_555 & \texttt{C}\_852 \\ \midrule
SCoNE-VD  & $0.768$  & $0.741$  & $0.753$  & $0.839$  & $0.747$  & $0.655$  \\
SCoNE-1NN & $0.848$  & $0.861$  & $0.873$  & $0.882$  & $0.785$  & $0.817$  \\ \midrule
SCoNE     & $\textbf{0.918}$  & $\textbf{0.941}$  & $\textbf{0.933}$  & $\textbf{0.967}$  & $\textbf{0.927}$  & $\textbf{0.945}$  \\ \bottomrule
\end{tabular}}
\caption{Ablation analysis of average AUC results on two real-world multi-view datasets, \texttt{AWA} and \texttt{Caltech}, using the same notation as in Table~\ref{tab:mainall}.}
\label{tab:ablation}
\end{table}


Experiments are performed on the largest $2,5,8$ and $10$ social networks in \texttt{Facebook}, as shown in Table~\ref{tab:facebook}. It is cast as a multi-view problem with $V=3$, and each view has randomly retained 70\% of its connectivity relationships and contains class anomalies with an anomaly ratio of 5\%. Our proposed method also achieve the best result on this task. When the number of social networks is higher than five, SCoNE outperforms LDSCEN, which is due to the increased likelihood of LDSCEN incorrectly categorizing instances into their respective social networks. Since SCoNE focuses on the consistent neighborhoods across all views, it is thereby more robust as the number of networks increases. 

\begin{table}[]
\centering
\resizebox{0.98\linewidth}{!}{
\begin{tabular}{l|cccc}
\toprule
        & {Facebook}-2  & {Facebook}-5 & {Facebook}-8 & {Facebook}-10 \\ \midrule
iForest & $0.742_{\pm 0.016}$ & $0.725_{\pm 0.013}$ & $0.727_{\pm 0.015}$ & $0.703_{\pm 0.014}$ \\
MODDIS  & $0.763_{\pm 0.008}$ & $0.737_{\pm 0.011}$ & $0.688_{\pm 0.010}$ & $0.669_{\pm 0.010}$ \\
HBM     & $0.707_{\pm 0.023}$ & $0.721_{\pm 0.018}$ & $0.724_{\pm 0.021}$ & $0.714_{\pm 0.019}$ \\
NCMOD   & $0.809_{\pm 0.011}$ & $0.774_{\pm 0.013}$ & $0.769_{\pm 0.014}$ & $0.723_{\pm 0.016}$ \\
ECMOD   & $0.856_{\pm 0.014}$ & $0.817_{\pm 0.019}$ & $0.803_{\pm 0.021}$ & $0.781_{\pm 0.024}$ \\
SRLSP   & $0.977_{\pm 0.003}$ & $0.968_{\pm 0.006}$ & $0.959_{\pm 0.009}$ & $0.954_{\pm 0.009}$ \\
MODGD   & $0.954_{\pm 0.002}$ & $0.949_{\pm 0.004}$ & $0.938_{\pm 0.005}$ & $0.931_{\pm 0.004}$ \\
IAMOD   & $0.981_{\pm 0.003}$ & $0.957_{\pm 0.003}$ & $0.951_{\pm 0.005}$ & $0.947_{\pm 0.006}$ \\ \midrule
LDSCEN  & $\textbf{0.991}_{\pm 0.002}$ & $0.972_{\pm 0.006}$ & $0.962_{\pm 0.007}$ & $0.955_{\pm 0.007}$ \\ 
SCoNE     & $0.982_{\pm 0.003}$ & $\textbf{0.973}_{\pm 0.003}$ & $\textbf{0.965}_{\pm 0.004}$ & $\textbf{0.960}_{\pm 0.004}$ \\ \bottomrule
\end{tabular}}
\caption{The mean AUC ± 2*SE results on \texttt{Facebook}.}
\label{tab:facebook}
\end{table}

\section{Conclusion}
\label{sec:con}

We introduce a paradigm shift in multi-view anomaly detection by moving from indirect, learning-based approaches to a direct representation of consistent neighborhoods using the multi-view instances themselves. This core insight enables SCoNE, an effective and efficient multi-view anomaly detector built upon adaptive spherical neighborhoods.
The reason for using spherical neighborhoods is due to their unique data-dependent properties, which are essential for adapting to varied data densities and producing robust consistency. We demonstrate that the consistent spherical neighborhood representation derived from multi-view instances enables it to effectively identify consistent neighbors across views on density difference data. Our theoretical analysis and empirical evaluations establish SCoNE’s superior detection accuracy and linear time complexity, making it highly suitable for large-scale multi-view anomaly detection task. Moreover, exploring the application of multi-view anomaly detection methods in industrial data, such as 3D anomaly data, is a potential future research direction.

\newpage
\section*{Acknowledgements}
 This work is supported by National Natural Science Foundation of China (Grant No. W2531050).

\bibliography{aaai2026}

\newpage

\appendix

\end{document}